\documentclass[runningheads]{llncs}

\usepackage{amsmath}
\usepackage{amssymb} 
\usepackage[T1]{fontenc}
\usepackage{newtxtext}
\usepackage[varvw]{newtxmath}
\usepackage{bm}

\usepackage{graphicx}
\usepackage{placeins}
\usepackage{adjustbox}
\usepackage{booktabs}
\usepackage{multirow}
\usepackage[table]{xcolor}
\newif\ifRevisionColor
\RevisionColorfalse
\newcommand{\revision}[1]{{\ifRevisionColor\color{red}\fi #1}}
\usepackage[font=footnotesize]{subcaption}
\usepackage{tabularx}

\usepackage{algorithm}
\usepackage{algpseudocode}

\usepackage[hidelinks]{hyperref}

\usepackage{cleveref}
\crefname{figure}{Fig.}{Figs.}  
\crefformat{equation}{(#2#1#3)}
\crefformat{appendix}{#2#1#3}
\Crefformat{appendix}{#2#1#3}  

\definecolor{myblue}{gray}{0.93} 
\definecolor{mypink}{gray}{0.80} 
\definecolor{deepteal}{RGB}{0, 80, 80}
\definecolor{deepblue}{RGB}{20, 60, 140}
\definecolor{deeppurple}{RGB}{90, 40, 130}

\newcommand{\e}[1]{\cellcolor{black!#1}}
\newcommand{\s}[1]{\cellcolor{black!#1}}

\title{Projection-Free Evolution Strategies for Continuous Prompt Search}
\titlerunning{Projection-Free Evolution Strategies for Prompt Search}
\author{Yu Cai \and Canxi Huang \and Xiaoyu He\thanks{Corresponding author.}}
\authorrunning{Y. Cai et al.}
\institute{School of Software Engineering, Sun Yat-sen University, Zhuhai, China\\
\email{caiy86@mail2.sysu.edu.cn, huangcx23@mail2.sysu.edu.cn, hexy73@mail.sysu.edu.cn}}

\begin{document}
\raggedbottom

\maketitle

\begin{abstract}
Continuous prompt search offers a computationally efficient alternative to conventional parameter tuning in natural language processing tasks. Nevertheless, its practical effectiveness can be significantly hindered by the black-box nature and the inherent high-dimensionality of the objective landscapes. 
Existing methods typically mitigate these challenges by restricting the search to a randomly projected low-dimensional subspace. 
However, the effectiveness and underlying motivation of the projection mechanism remain ambiguous.
In this paper, we first empirically demonstrate that despite the prompt space possessing a low-dimensional structure, random projections fail to adequately capture this essential structure.
Motivated by this finding, we propose a projection-free prompt search method based on evolutionary strategies. 
By directly optimizing in the full prompt space with an adaptation mechanism calibrated to the intrinsic dimension,
our method achieves competitive search capabilities without additional computational overhead.
Furthermore, to bridge the generalization gap in few-shot scenarios, we introduce a confidence-based
regularization mechanism that systematically enhances the model's confidence in the target verbalizers. 
Experimental results on 
seven natural language understanding tasks from the GLUE benchmark
demonstrate that our proposed approach outperforms existing baselines.
\end{abstract}
 
\section{Introduction}
\revision{Transformer-based pre-trained language models (PLMs) have attracted attention for natural language processing applications, including text rephrasing and paraphrasing}~\cite{qiu2020pre}.
Recent studies have suggested that the performance of PLMs largely depends on the model size, and it is expected that, to approach human-level performance, a PLM must involve hundreds of billions of parameters~\cite{zhao2023survey}. 
\revision{Even tuning a small number of parameters can be prohibitively expensive because backpropagation requires substantial activation memory.}
\revision{Modern PLMs such as GPT-4~\cite{achiam2023gpt} are also commonly offered as services, making direct gradient-based parameter updates intractable.}

\revision{Prompt search offers an efficient alternative to parameter tuning.}
Prompt here refers to additional information, e.g., a common prefix sentence, which is fed into a PLM along with original input sentences. Prompt search treats PLMs as a black-box and optimizes some metrics over prompts rather than the model parameters. We focus on continuous prompt, where the prompt is a token sequence and the decision space of prompt search is the product of individual embedding spaces of all involved tokens (or the ``prompt space'' for short).

\revision{The black-box optimization difficulty of prompt search lies in its high dimensionality.}
For example, RoBERTa-Large~\cite{liu2019roberta} has an embedding space of dimension 1,024, so searching for merely 50 prompt tokens leads to a problem of dimension 51,200.
\revision{Black-box optimization convergence slows with dimension and can be extremely slow on high-dimensional landscapes}~\cite{nesterov2017random}.
Furthermore, modern black-box optimization algorithms, including both metaheuristics and those based on objective/gradient estimation, are seldom used when the problem has more than 10,000 dimensions due to the high computational cost~\cite{larson2019derivative}. 
Therefore, applying these methods directly to prompt search could be infeasible.

\revision{A representative approach is Sun \textit{et al.}'s Black-Box Tuning (BBT)}~\cite{sun2022black}. 
BBT is a metaheuristic method based on projection.
\revision{It assumes that high-quality prompts can be found in a low-dimensional affine subspace with a fixed basis.}
\revision{Any black-box optimizer can then operate in the reduced space, addressing the high-dimensional challenge.} 
Sun \textit{et al.} suggested pre-defining the basis of the subspace via random sampling and then solving the space-reduced problem with Covariance Matrix Adaptation Evolution Strategy (CMA-ES)~\cite{hansen2003reducing}, a well-established metaheuristic based on second-order information approximation. 
\revision{The method performs well on several language understanding tasks and can match standard full-parameter tuning}~\cite{sun2022bbtv2}.

\revision{Despite BBT's success, its projection mechanism remains poorly understood.}
For example, it assumes that the prompt space
is intrinsically low-dimensional.
\revision{Studies on low-dimensional PLM reparameterization partly support this assumption~\cite{aghajanyan2020intrinsic}, but whether it holds in prompt space remains unclear. This raises our first research question.}

\textit{Q1. Does the prompt space of PLMs really have a low-dimensional structure?} 

\revision{On seven GLUE tasks, we confirm that the prompt space has an intrinsic dimension of only a few hundred.}
In particular, we show empirically that there exists an upper bound on the estimated intrinsic dimension that does not increase consistently with the prompt length.This verifies the low-dimensional assumption adopted by BBT and similar work.

\revision{However, low intrinsic dimension does not imply a global subspace that captures high-quality prompts.} 
\revision{Our experiments instead show that low-dimensional prompt projection is ineffective.}
\revision{We therefore propose projection-free alternatives to BBT that exploit low intrinsic dimension, leading to our second research question.} 

\textit{Q2. How to efficiently perform prompt search without subspace projections?}

We provide a straightforward yet effective approach based on the evolution strategy (ES) family of metaheuristics: instead of explicitly projecting prompts, we leave the decision space untouched and calibrate the adaptation mechanism to the intrinsic dimension, enabling standard ES variants to efficiently perform prompt search.

\revision{Our final question concerns improving generalization in projection-free search:}

\textit{Q3. How to regularize the objective of prompt search to promote generalization?}

\revision{We argue that low global confidence in the target verbalizers can increase the risk of overfitting in full-space prompt search. This motivates a confidence-based regularization term that encourages greater probability mass on the verbalizer set, with the aim of improving generalization.}

\revision{We address these questions with a simple projection-free framework for continuous prompt search that outperforms existing baselines.}
The key findings are summarized below:
\begin{itemize}
\item We confirm the low-dimensional nature of the prompt search problem and quantify the intrinsic dimension.
\item We show that standard ESs can efficiently perform prompt search in the full space by recalibrating the adaptation mechanism.
\item We propose confidence-based regularization to enhance generalization for projection-free prompt search.
\end{itemize}

\section{Preliminaries}
We formulate the prompt search problem and review BBT.

\subsection{Problem Definition}
Following \cite{sun2022black}, we formulate the continuous prompt of length $l$ as a flattened vector $x \in \mathbb{R}^{d}$, where $d = l \cdot e$ represents the full prompt space dimension and $e$ denotes the embedding size of the PLM $\mathcal{M}$.

The quality of a prompt $x\in\mathbb{R}^{d}$ is evaluated via querying $\mathcal{M}$ on some data distribution conditioned on $x$. The task of prompt search can therefore be stated as:
\begin{equation}\label{eq:ps}
\min_{x\in\mathbb{R}^{d}} f_{PS}(x) = \mathbb{E}_{(y,z)\sim\mathcal{D}} [\ell(\mathcal{M}(x \oplus y), z)] 
\end{equation}
where $\ell$ is a loss function, $\mathcal{D}$ is a data distribution, $(y, z)$ denotes a data sample containing an input sentence $y$ and its corresponding label $z$, and $\oplus$ denotes augmenting an input sentence with the prompt. 
The operation $x \oplus y$ transforms the input and prompt into the model's embedding space through the following four steps: 
1) preparing a template with specific placeholders; 
2) placing the input $y$ into the corresponding placeholder locations; 
3) computing the embedding vector of the resultant sentence; 
4) concatenating the continuous prompt vector $x$ with the embedding vector obtained.

We adapt $\mathcal{M}$ for classification by evaluating log-likelihoods at the specific target position: the mask location for masked language models, or the first token following the prefix for causal language models. 
Let $\mathcal{M}(x \oplus y) \in \mathbb{R}^{|\mathcal{V}|}$ denote the logits over the vocabulary $\mathcal{V}$. 
The label $z$ maps to a verbalizer subset $\mathcal{W} \subset \mathcal{V}$\footnote{\revision{Our task templates and label verbalizers are based on the prompting setup of BDPL~\cite{diao2022black}, with adaptations for the respective model architectures. See \Cref{appx:2} for the task-specific templates and verbalizers.}}. 
We minimize the cross-entropy loss:
\begin{equation}\label{eq:ce}
\ell_{CE}(\bm{a}, z) = - \sum_{v \in \mathcal{W}} \mathbb{I}\{v=z\} \log \left( \frac{\exp(a_v)}{\sum_{u \in \mathcal{W}} \exp(a_u)} \right)
\end{equation}
where $\bm{a} \in \mathbb{R}^{|\mathcal{V}|}$ represents the logits over the vocabulary and $a_v$ corresponds to token $v$.
Thus minimizing $\ell(\mathcal{M}(x\oplus y),z)$ increases the likelihood of producing the correct label $z$ given an input $y$ and the prompt~$x$.

\subsection{The BBT Method}\label{sec:bbt}
BBT~\cite{sun2022black} is an evolutionary approach to solving problem \cref{eq:ps}.
Its underlying algorithm is CMA-ES, a de facto standard metaheuristic for real-valued black-box optimization.
CMA-ES explores the prompt space via sampling a multivariate Gaussian distribution, and the distribution is updated iteratively using landscape curvature information estimated via zeroth-order oracles.
Hence it requires maintaining a covariance matrix of $d \times d$, leading to a quadratic complexity.
BBT reduces the complexity by parameterizing prompts via a $\tilde{d}$-dimensional subspace and optimizing the following objective:
\begin{equation}\label{eq:bbt}
    \min_{x\in \mathbb{R}^{\tilde d}} f_\text{BBT}(x;A,x_\text{init})= f_\text{PS}(x_\text{init}+Ax)
\end{equation}
where $f_\text{PS}$ is given in \cref{eq:ps}. 
Here, $x_\text{init} \in \mathbb{R}^d$ and $A \in \mathbb{R}^{d\times {\tilde d}}$ denote the initial prompt (or anchor point) and the random projection matrix, respectively.
Both terms are predefined and fixed throughout the search process.
Hence, the decision space of \cref{eq:bbt} is a $\tilde{d}$-dimensional affine subspace with a fixed basis.
With a sufficiently small $\tilde{d}$, the algorithmic overhead of CMA-ES can be made negligible.

\section{Low Intrinsic Dimension of Prompt Space}
As mentioned above, BBT randomly generates the initial prompt $x_\text{init}$ and projection matrix $A$ and fixes them during the search. 
This implicitly assumes 1) the existence of a global subspace in which the prompt search problem can be solved approximately, and 2) such a subspace can be constructed via random projections.
We show in this section that these assumptions are not fully supported by empirical evidences.
On one hand, we do find the prompt space admits a manifold which is of low dimension.
On the other hand, random projection does not capture promising search regions, thereby not making problem-solving truly easier.
\subsection{Low-Dimensional Manifold of Prompt Space}
\label{sec:measureID}
We investigate empirically the manifold geometry of the prompt space and show that it admits a low intrinsic dimension (ID).
Our experiment involves a set of random prompts $\mathcal{X}$. Each prompt is uniformly sampled from the discrete prompt space $\mathcal{V}^l$ and then cast into the continuous prompt space $\mathbb{R}^d$.
To reflect the real hardness of prompt search, we explicitly evaluate the gradients of the objective \cref{eq:ps} at these prompts as they capture the directions along which the prompts can be improved\footnotemark. 
\footnotetext{Gradient access is utilized solely for this analysis. All prompt search methods in this work operate under a gradient-free setting.}
Let $\mathcal{G} = \{\nabla f_\text{PS}(x)|x\in\mathcal{X}\}$ denote the gradient set, and we expect $\mathcal{G}$ to form a low-dimensional manifold so prompts can be improved without exploring the full prompt space. 
We first normalize the gradients in $\mathcal{G}$ to have zero mean and unit variance, and then compute their pairwise cosine distances. 
The intrinsic dimension of the manifold spanned by $\mathcal{G}$ can now be estimated using the maximum likelihood estimation proposed in \cite{levina2004maximum} as:
\[
\tilde{d} = \frac{1}{|\mathcal{X}|}\sum_{x \in \mathcal{X}}\left(\frac{1}{k-1} \sum_{j=1}^{k-1} \log \frac{T_k(x)}{T_j(x)}\right)^{-1},
\]
where $T_k(x)$ denotes the distance of $\nabla f(x)$ to its $k$-th nearest neighbor in $\mathcal{G}$.
This metric captures the minimum number of embedding dimensions needed for preserving the local neighborhood structure in $\mathcal{G}$. 

General settings of the experiment are listed below.
We set $|\mathcal{X}|=5,000$.
We use RoBERTa‑Large as the backbone model~\cite{liu2019roberta}. 
We consider seven tasks from the GLUE benchmark \cite{wang2018glue}, namely MRPC, MNLI, QQP, SST-2, RTE, CoLA, and QNLI.
The experiment follows a 16-shot training setting. 
To ensure a thorough analysis, we evaluate the intrinsic dimension across a comprehensive range of prompt lengths $l$ and neighborhood sizes $k$.

\begin{figure}[!b]
  \centering
  \small 
  \includegraphics[width=0.95\columnwidth]{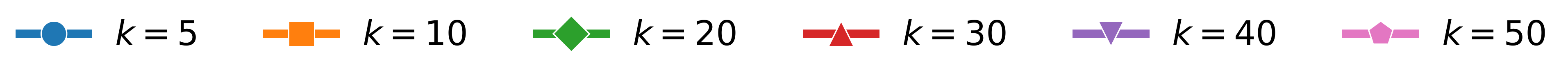} \\
  \captionsetup[subfigure]{skip=0.5pt} 

  \subcaptionbox{ MRPC}{\includegraphics[width=0.25\columnwidth]{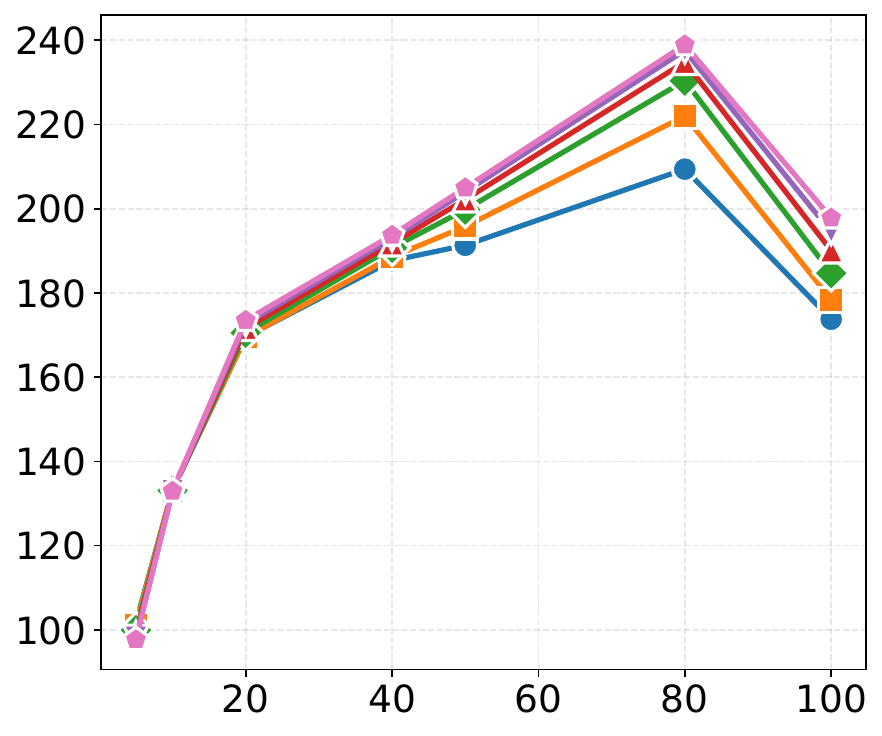}}%
  \subcaptionbox{ MNLI}{\includegraphics[width=0.25\columnwidth]{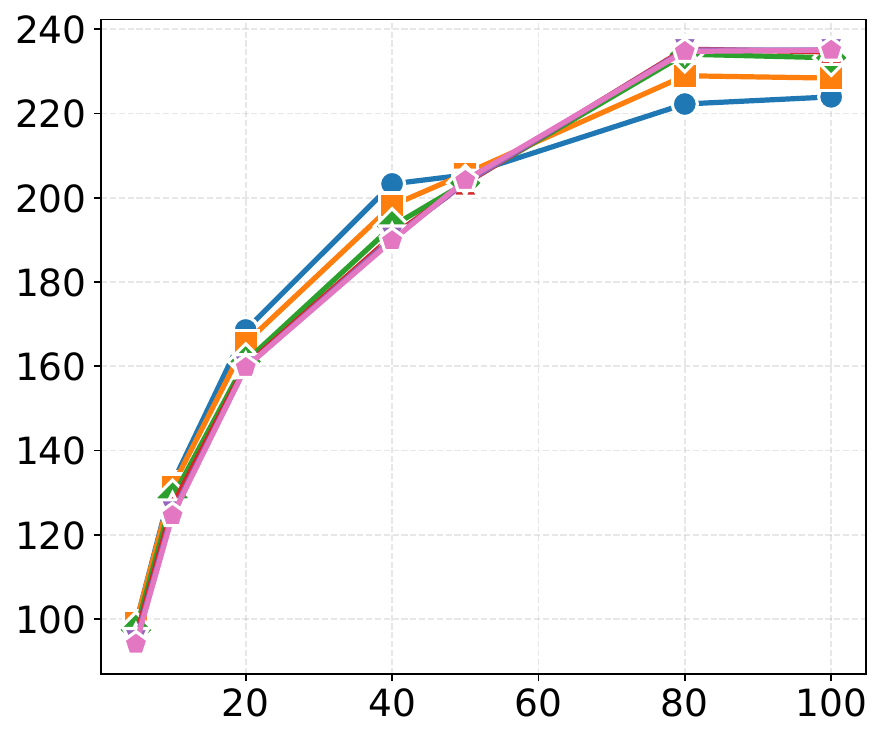}}%
  \subcaptionbox{QQP}{\includegraphics[width=0.25\columnwidth]{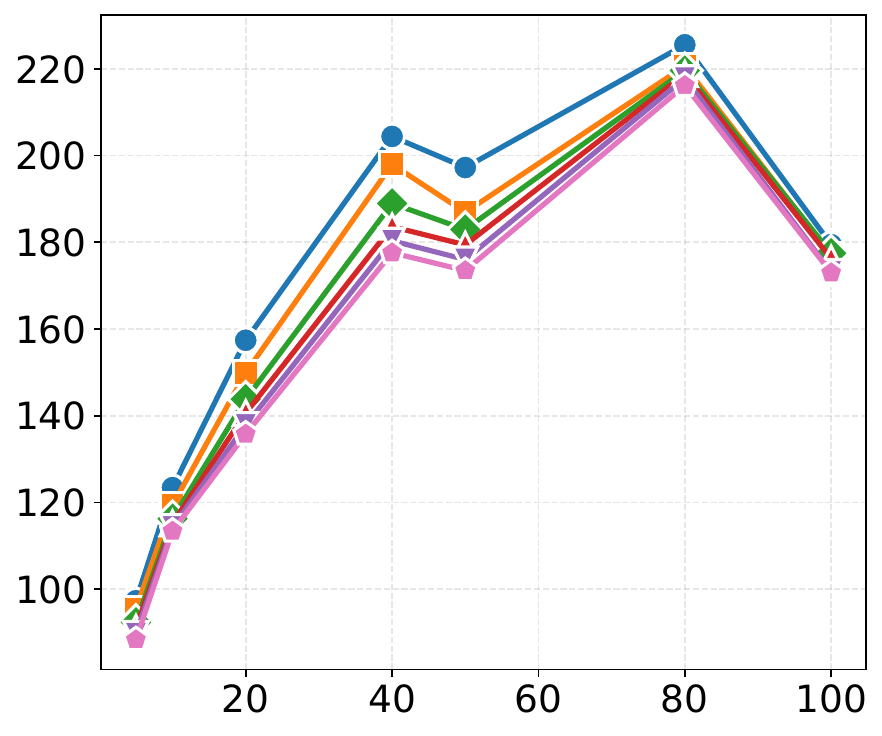}}%
  \subcaptionbox{SST-2}{\includegraphics[width=0.25\columnwidth]{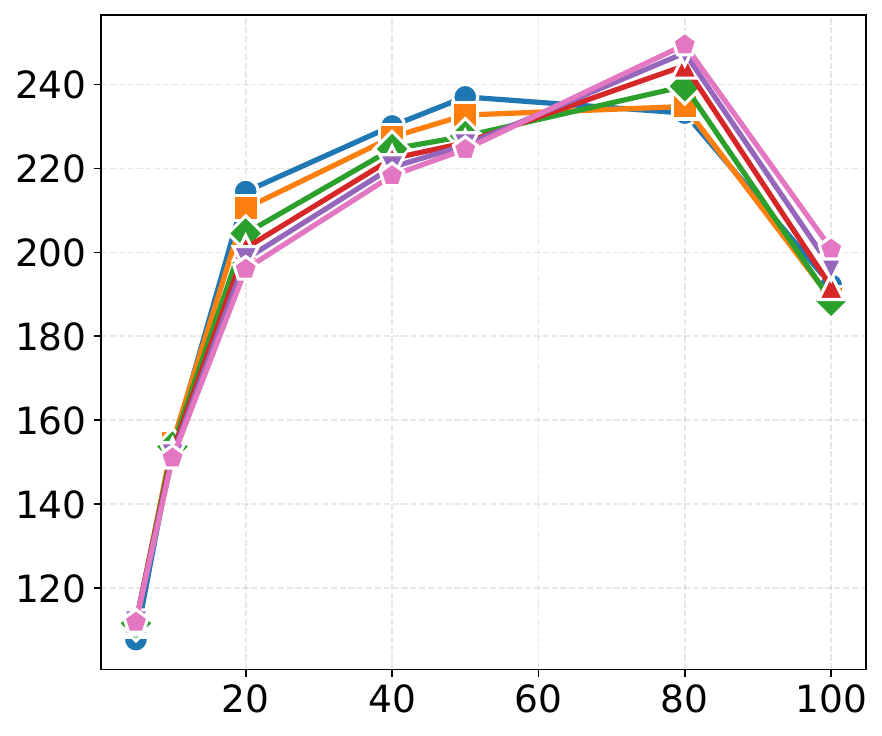}}

  \hspace*{\fill}
  \subcaptionbox{RTE}{\includegraphics[width=0.25\columnwidth]{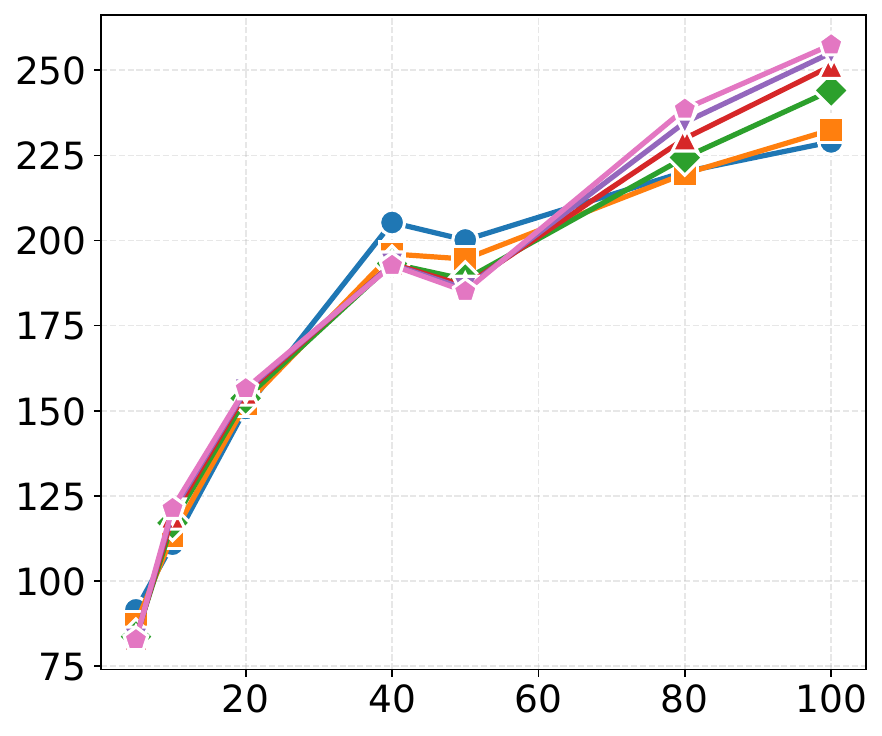}}\hfill
  \subcaptionbox{CoLA}{\includegraphics[width=0.25\columnwidth]{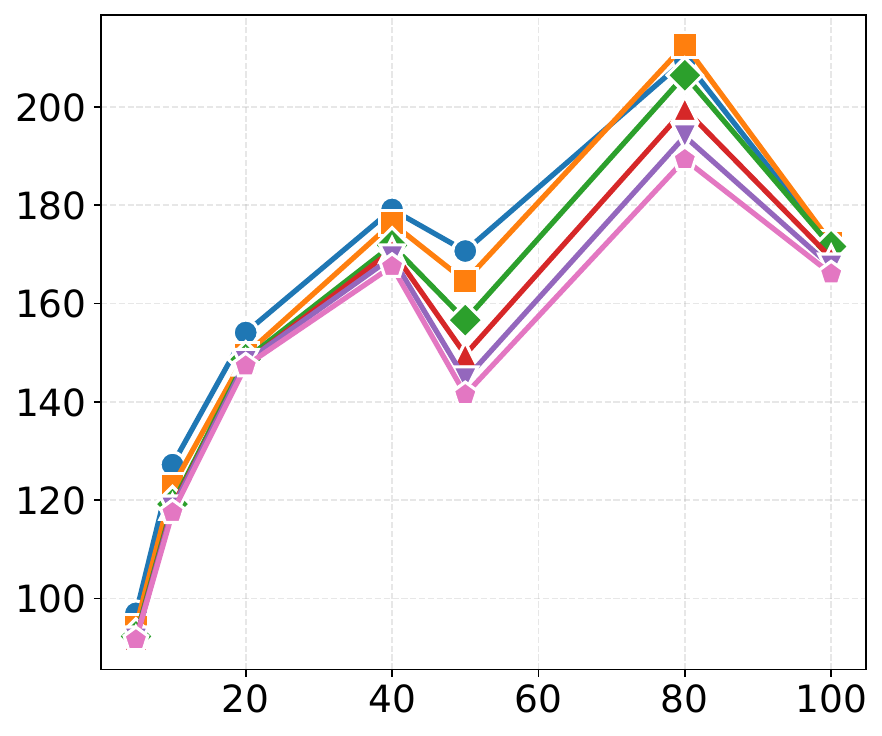}}\hfill
  \subcaptionbox{QNLI}{\includegraphics[width=0.25\columnwidth]{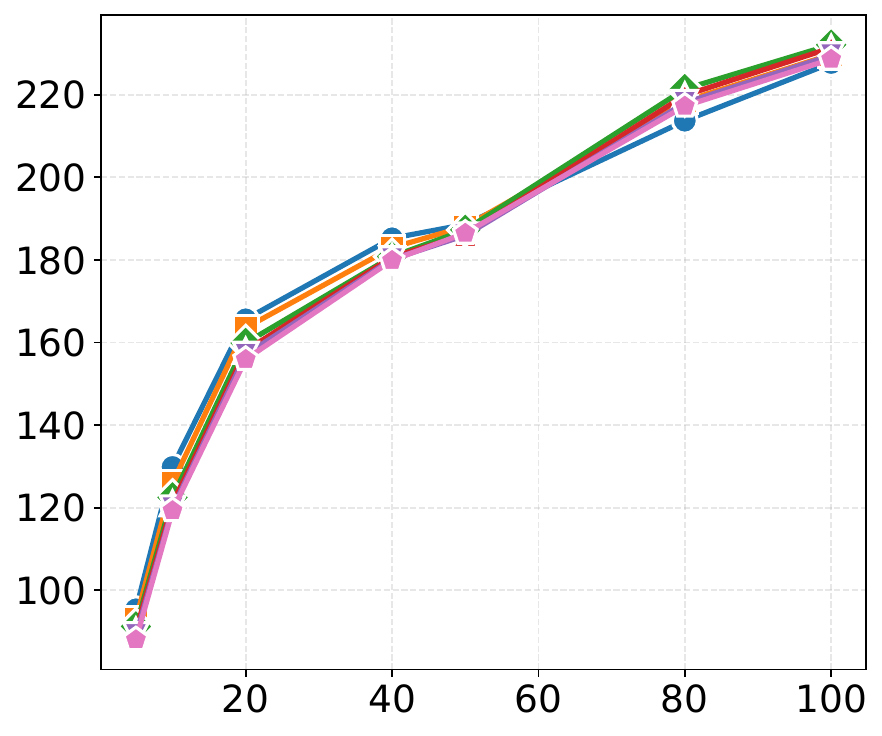}}
  \hspace*{\fill}

  \caption{Intrinsic dimension estimation on seven GLUE tasks. We report the estimated dimension (y-axis) across continuous prompt lengths $l \in \{5, 10, 20, 40, 50, 80, 100\}$ (x-axis). Different colors correspond to the neighborhood size $k \in \{5, 10, 20, 30, 40, 50\}$ used in the maximum likelihood estimation.}  
  \label{fig:id}
\end{figure}
\Cref{fig:id} plots the estimated intrinsic dimension $\hat{d}$ across varying prompt lengths. 
The estimation proves robust to neighborhood size $k$ and, crucially, does not increase proportionally with the prompt length. 
Instead, it consistently exhibits an empirical upper bound of approximately $250$ across tasks. 
This confirms that the intrinsic dimension is orders of magnitude lower than the ambient dimension $d$, affirmatively answering Q1.  
This suggests prompt optimization is feasible within a low-dimensional manifold, independent of the full prompt space size.
See \Cref{sec:sensitive} for sensitivity analysis.

\subsection{The Limitations of Random Projections}
\label{sec:subspace_issue}
\noindent\revision{\textit{General Experimental Setup.}\enspace We consider seven GLUE tasks~\cite{wang2018glue}: MRPC, MNLI, QQP, SST-2, RTE, CoLA, and QNLI. For each task, the training and validation sets each contain 16 randomly sampled examples per class, and model selection uses validation performance. We report MCC for CoLA, F1 for MRPC and QQP, and accuracy for the remaining tasks, using the matched validation split for MNLI. Our experimental results are reported as means and standard deviations over five runs with seeds $\{0,1,42,43,100\}$. Experiments run on NVIDIA RTX 4090 GPUs. Unless otherwise stated, the continuous prompt lengths are $l=50$ for RoBERTa-Large and $l=10$ for OPT-6.7B, and each optimization run uses 5,000 function evaluations (FEs).}
\revision{\paragraph{Configuration for the Projection Comparison.} We use RoBERTa-Large and shorten the prompt to $l=5$ ($d=5{,}120$) to make standard CMA-ES computationally feasible in the full space. Both the full-space and subspace settings use standard CMA-ES and the same initial prompt, with the subspace dimension fixed at $\tilde{d}=500$. Both settings receive a budget of 10,000 FEs, deliberately allowing headroom to provide sufficient opportunity for convergence.}
\begin{table}[!t]
  \centering
  \caption{Comparison of training loss on the evaluated tasks. $f_{\mathrm{PS}}(x_*)$ denotes the approximate optimal value in the full prompt space, while $f_{\mathrm{BBT}}(y_*)$ corresponds to that in the random subspace.}
  \label{table:subspace-projection}
  \small
  \setlength{\tabcolsep}{10pt}
  \begin{tabular}{lcccc}
  \toprule
  & \multicolumn{1}{c}{$f_{\mathrm{PS}}(x_*)$} & \multicolumn{1}{c}{$f_{\mathrm{BBT}}(y_*)$} & \multicolumn{1}{c}{$\gamma_\text{OP}$} & \multicolumn{1}{c}{$\gamma_\text{PI}$} \\
  \midrule
  MRPC  & 0.0250 & 0.0413 & 1.0072 & 1.4175 \\
  MNLI  & 0.2479 & 0.2738 & 0.9919 & 1.4001 \\
  QQP   & 0.0073 & 0.0310 & 0.9665 & 1.3856 \\
  SST-2  & 0.0051 & 0.0100 & 0.9590 & 1.3843 \\
  RTE   & 0.0611 & 0.0845 & 0.9632 & 1.3864 \\
  CoLA  & 0.0416 & 0.0698 & 0.9698 & 1.3878 \\
  QNLI  & 0.0156 & 0.0202 & 0.9886 & 1.4055 \\
  \bottomrule
  \end{tabular}
\end{table}
\revision{Even with a low-dimensional manifold, prompts may not be easily improved in a randomly generated, fixed subspace. We compare solving the space-reduced problem \cref{eq:bbt} in $\mathbb{R}^{\tilde d}$ with solving the original problem \cref{eq:ps} directly in $\mathbb{R}^d$. We show that the former neither reduces the computational effort nor improves solution quality. Let $x_*\in\mathbb{R}^d$ be the reference optimal prompt obtained by full-space CMA-ES from $x_\text{init}$, and let $y_*\in\mathbb{R}^{\tilde d}$ be the approximate solution obtained by BBT from a zero vector. The corresponding full-space prompt is $Ay_*+x_\text{init}$, so both searches start from the same prompt.}

\revision{Besides comparing objective values, we use two distance-based metrics to assess optimization difficulty: relative optimization progress $\gamma_\text{OP}$ and relative post-hoc improvement $\gamma_\text{PI}$, defined as}
\[
    \gamma_\text{OP} := \frac{\|(Ay_*+x_\text{init}) - x_\text{init}\|}{\|x_* - x_\text{init}\|},
    \qquad
    \gamma_\text{PI} = \frac{\|x_* - (Ay_*+x_\text{init})\|}{\|x_* - x_\text{init}\|}.
\]
\revision{Both metrics use the full-space displacement $\|x_*-x_\text{init}\|$ as the reference. For $\gamma_\text{OP}$, optimization progress is measured by the Euclidean distance from the initial prompt to the obtained solution; a lower ratio indicates greater efficiency of subspace search relative to full-space search. For $\gamma_\text{PI}$, the numerator quantifies the additional effort to close the gap between the subspace and full-space solutions; a value above 1 means that post-hoc improvement of the subspace solution could be more expensive than solving \cref{eq:ps} directly.}

\revision{\Cref{table:subspace-projection} shows that full-space search yields lower training loss on all seven tasks, demonstrating that superior prompts can be obtained in the full space. Meanwhile, $\gamma_\text{OP}$ remains close to 1, implying that subspace search does not effectively shorten the optimization trajectory. Projection primarily circumvents the prohibitive cost of high-dimensional second-order methods, which can alternatively be addressed using simpler, scalable algorithms. Finally, $\gamma_\text{PI}>1$ on every task suggests that post-hoc improvement of the subspace solution could cost more than direct full-space optimization. Together, these results indicate that random subspace projection limits achievable solution quality, creating a gap not effectively bridged by merely extending the search budget.}

\section{Projection-Free Evolution Strategies}\label{sec:id}
Since random subspace projections offer limited efficiency gains while significantly hindering the search for optimal prompts, as shown in \Cref{sec:subspace_issue}, we propose a projection-free method to directly search within the full prompt space. 
We employ ES variants scaling linearly with the dimension, avoiding the quadratic complexity of covariance-based methods. 
However, these algorithms typically suffer from slow convergence in high-dimensional landscapes. 
To address this limitation and answer Q2, we introduce an ID-aware adaptation mechanism designed to accelerate optimization.

\subsection{ID-Aware Adaptation Mechanism}
We employ two ES variants, the (1+1)-ES and the SaES (collectively referred to as standard ES in their original forms, and ES-ID when equipped with our adaptation), to perform this projection-free search.
Both methods operate via isotropic Gaussian mutation, parameterized by a mean vector $x^{[t]}$ and a global step-size $\sigma^{[t]}$. 
The (1+1)-ES is a single-individual elitist method that adjusts $\sigma^{[t]}$ based on the 1/5 success rule~\cite{akimoto2022global}, while SaES utilizes a population-based self-adaptation mechanism where step-sizes are co-evolved. 
Detailed workflows are outlined in \Cref{alg:1+1ES,alg:SaES}.

Crucially, the step-size adaptation in both algorithms is governed by a damping factor $\tau$. 
Standard theoretical guidelines recommend scaling $\tau$ with the ambient dimension ($\tau = \sqrt{2d}$). 
However, as verified in \Cref{sec:measureID}, the prompt space admits a low-dimensional structure. 
Consequently, adhering to this conventional configuration results in excessively slow adaptation, where the step-size evolution is overly dampened by the high ambient dimensionality, hindering efficient convergence.

To address this, we propose the ID-aware adaptation mechanism that calibrates $\tau$ to the configured intrinsic dimension $\tilde{d}$ rather than $d$. 
While $\tau=\sqrt{2d}$ is standard for general black-box optimization, our proposed $\tau=\sqrt{2\tilde{d}}$ provides a distinct hyperparameter recommendation tailored for tasks exhibiting such low-dimensional structure, like prompt search.
\begin{algorithm}[htb]
  \footnotesize
  \caption{\colorbox{myblue}{Standard (1+1)-ES} and \colorbox{mypink}{(1+1)-ES-ID}}
  \label{alg:1+1ES}
  \begin{algorithmic}[1]
      \Require objective function $f:\mathbb{R}^d\to \mathbb{R}$, initial step-size $\sigma^{[0]} \in \mathbb{R}_+$, initial solution $x^{[0]} \in \mathbb{R}^d$, \colorbox{mypink}{intrinsic dimensions $\tilde{d}$}
      \State Setup damping factor: \colorbox{myblue}{$\tau=\sqrt{2d}$}  or  \colorbox{mypink}{$\tau=\sqrt{2\tilde{d}}$} 
  
      \For{$t = 0,1,\ldots$}
          \State $u^{[t]} \sim \mathcal{N}(0, I_d)$ \hfill \Comment{Mutation}
          \State $s \gets \mathbb{I}\{f(x^{[t]} + \sigma^{[t]} u^{[t]}) \leq f(x^{[t]})\}$ \hfill \Comment{Success indicator}
          \State $x^{[t+1]} = x^{[t]} + \sigma^{[t]} u^{[t]} \cdot s$ \hfill \Comment{Descent}
          \State $\sigma^{[t+1]} = \sigma^{[t]} \cdot \exp\left(\frac{1}{\tau}(s - \frac{1}{5})\right)$ \hfill \Comment{Step-size adaptation}
      \EndFor
\end{algorithmic}
\end{algorithm}
\begin{algorithm}[htb]  
\footnotesize
\caption{\colorbox{myblue}{Standard SaES} and \colorbox{mypink}{SaES-ID}}
\label{alg:SaES}
\begin{algorithmic}[1]  
    \Require objective function $f:\mathbb{R}^d \to \mathbb{R}$, initial step-size $\sigma^{[0]} \in \mathbb{R}_+$, initial solution $x^{[0]} \in \mathbb{R}^d$, population size $\lambda$, number of chosen solutions $\mu$, \colorbox{mypink}{intrinsic dimensions $\tilde{d}$}
    \State Setup damping factor: \colorbox{myblue}{$\tau=\sqrt{2d}$}  or  \colorbox{mypink}{$\tau=\sqrt{2\tilde{d}}$} 
    \For{$t = 0,1,\ldots$}
        \For{$i = 1,\ldots,\lambda$}
            \State $\delta_i^{[t]} \sim \mathcal{N}(0,1)$ \Comment{Step-size mutation}
            \State $\sigma_i^{[t]}=\sigma^{[t]}e^{\delta_i^{[t]}/\tau}$ \Comment{Trial step-size}
            \State $u_i^{[t]} \sim \mathcal{N}(0,I_d)$ \Comment{Solution mutation}
            \State $x_i^{[t]}=x^{[t]}+\sigma^{[t]}_i u_i^{[t]}$ \Comment{Trial solution}
        \EndFor
        \State Sort the solutions such that $f(x_{1:\lambda})\le\ldots\le f(x_{\lambda:\lambda})$
        \State  $x^{[t+1]}=\frac{1}{\mu} \sum^\mu_{i=1}x_{i:\lambda}^{[t]}$ \Comment{Recombination}
        \State  $\sigma^{[t+1]}=\frac{1}{\mu} \sum^\mu_{i=1}\sigma_{i:\lambda}^{[t]}$ \Comment{Step-size self-adaptation}
    \EndFor
\end{algorithmic}
\end{algorithm}
\subsection{Numerical Studies}
\begin{figure}[!b] 
  \centering
  \small
  \includegraphics[width=0.7\textwidth]{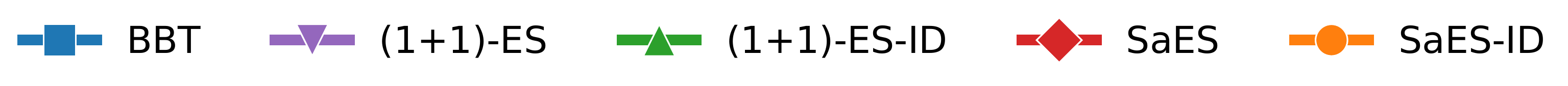}\\[6pt] 
  
  \captionsetup[subfigure]{justification=centering, font=footnotesize}
  \subcaptionbox{MRPC}{\includegraphics[width=0.235\textwidth]{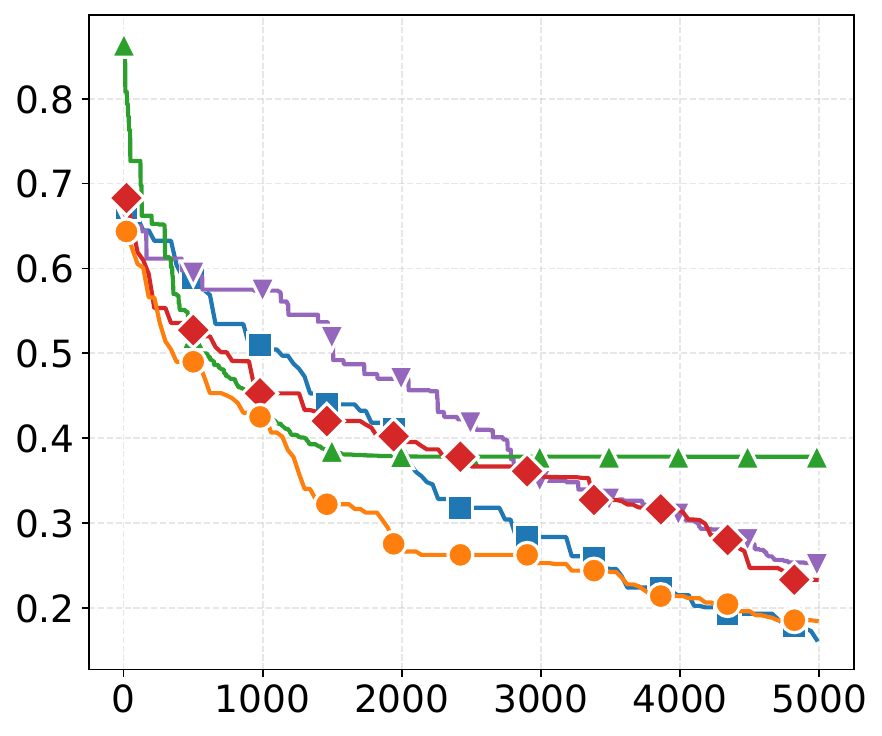}}\hfill
  \subcaptionbox{MNLI}{\includegraphics[width=0.235\textwidth]{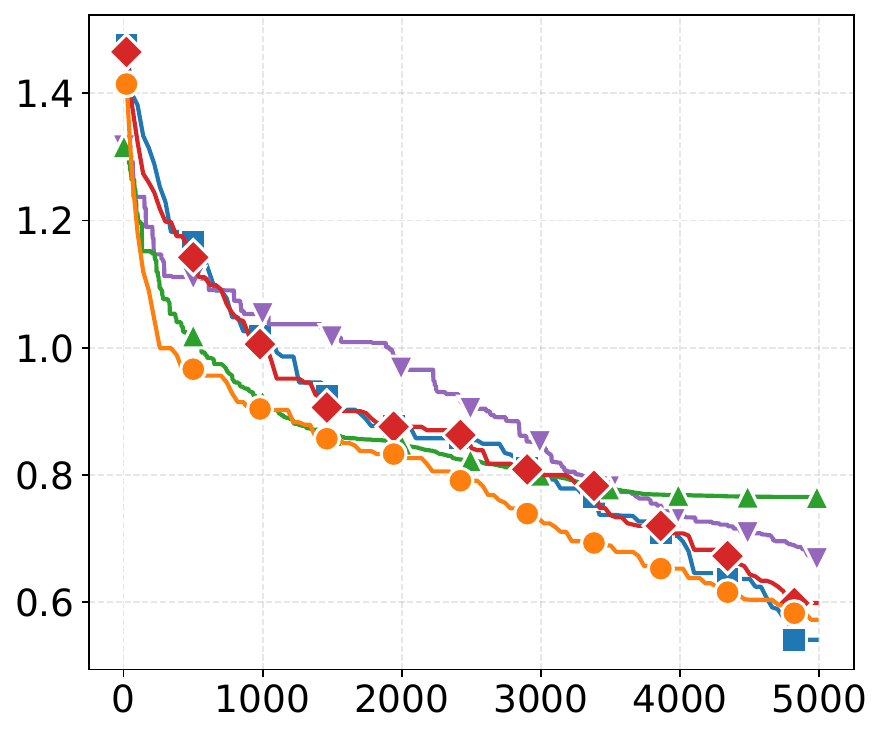}}\hfill
  \subcaptionbox{QQP}{\includegraphics[width=0.235\textwidth]{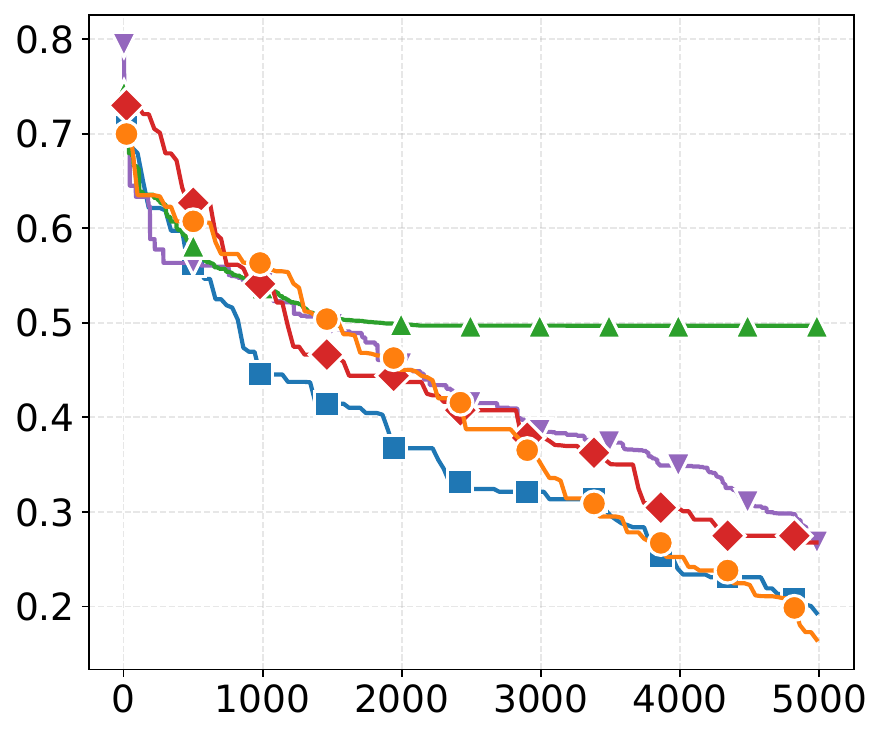}}\hfill
  \subcaptionbox{SST-2}{\includegraphics[width=0.235\textwidth]{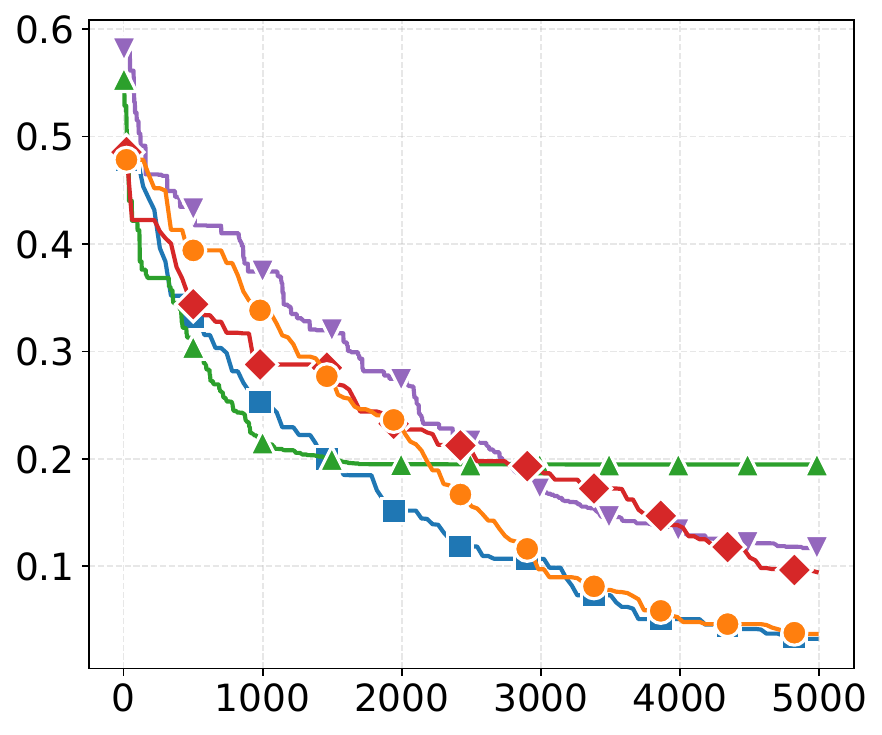}}
  
  \subcaptionbox{RTE}{\includegraphics[width=0.235\textwidth]{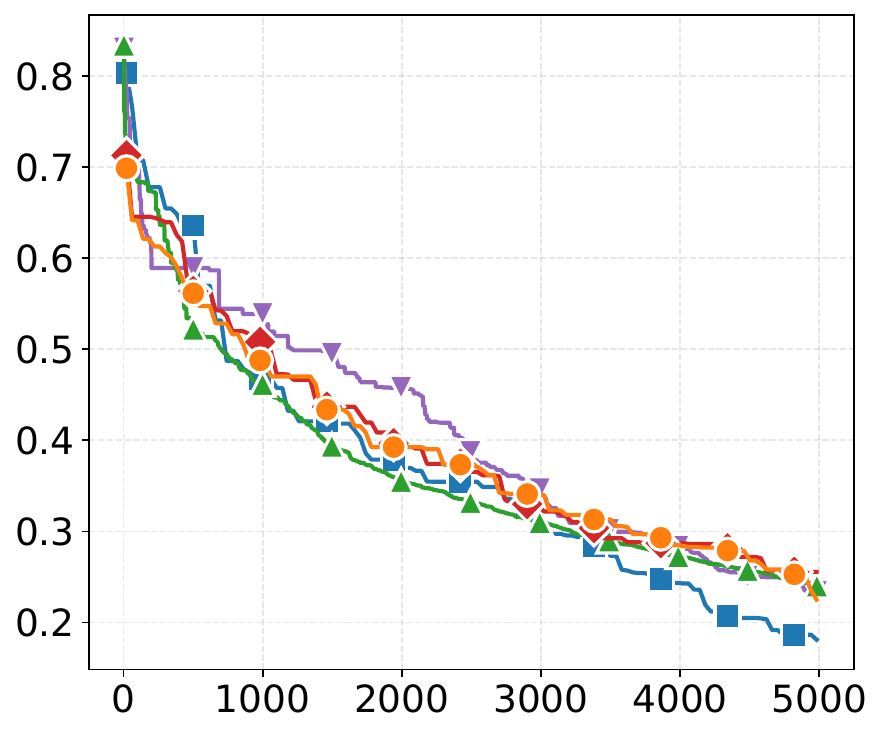}}\quad
  \subcaptionbox{CoLA}{\includegraphics[width=0.235\textwidth]{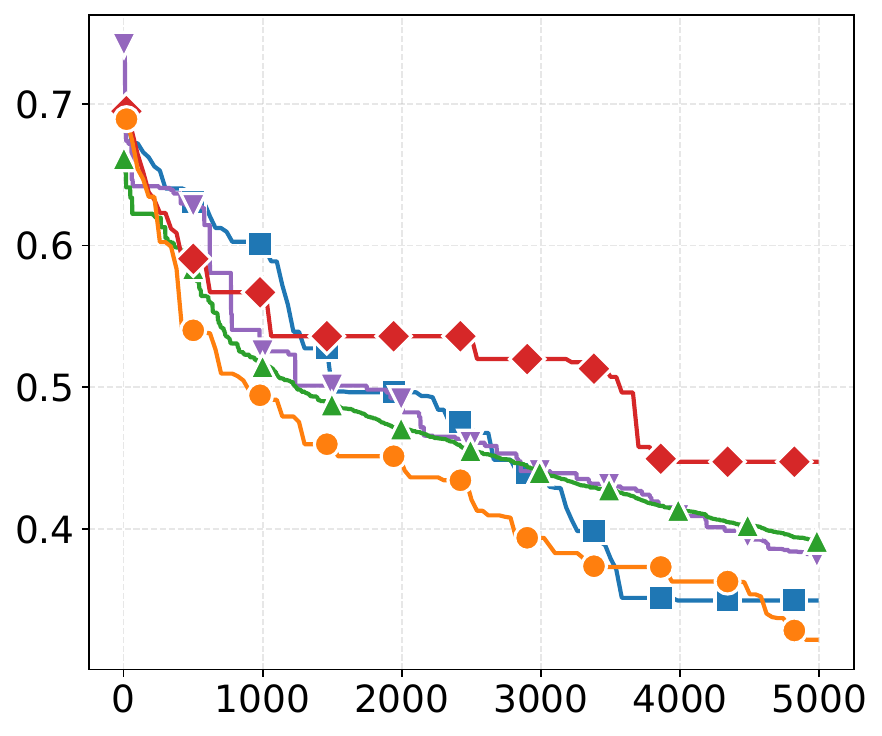}}\quad
  \subcaptionbox{QNLI}{\includegraphics[width=0.235\textwidth]{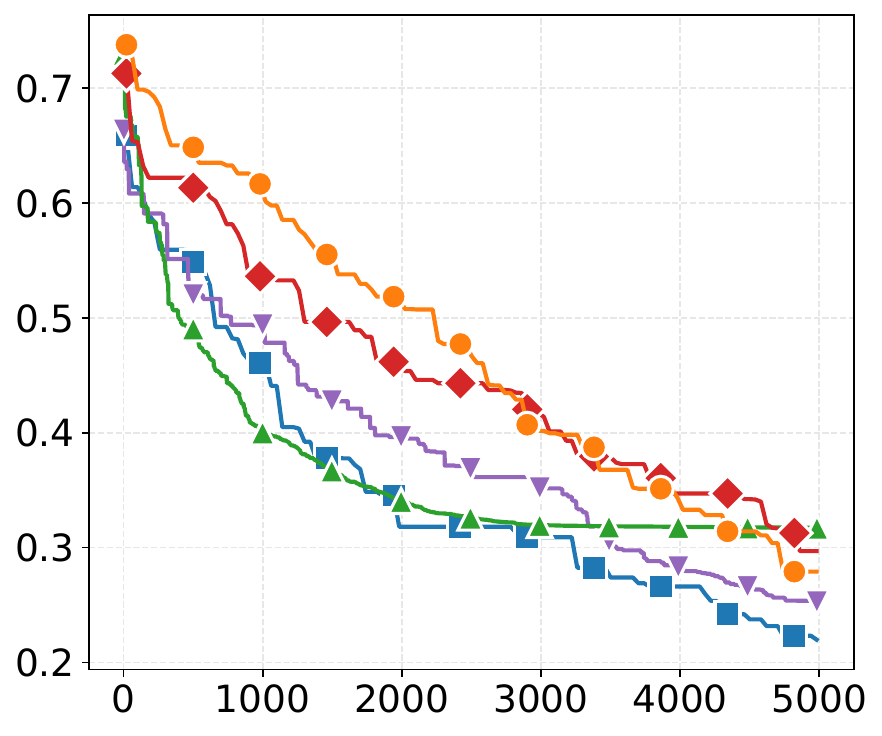}}
  \caption{Median performance on training data over 5 runs with  different random seeds. Algorithms include (1+1)-ES, (1+1)-ES-ID, SaES, SaES-ID with projection-free and BBT with  subspace projection. The x-axis denotes function evaluations, while the y-axis represents the loss function value.}\label{fig:ES-compare}
\end{figure}

We first verify the training efficiency by comparing ES-ID against BBT and standard ES. 
For fair comparison, we configure the intrinsic dimension $\tilde{d}=500$ for ES-ID to align with BBT's subspace dimensionality (see~\Cref{sec:sensitive} for an in-depth analysis of this configuration).
\revision{As illustrated in \Cref{fig:ES-compare}, SaES-ID achieves a training loss comparable to that of BBT within the limited budget of 5,000 function evaluations.} 
\revision{This confirms that our ID-aware adaptation effectively enables direct full-space search to match the performance of subspace-based approaches on most tasks.} 
In contrast, standard SaES often struggles to converge within this budget, as seen in CoLA. 
Although the single-individual (1+1)-ES-ID exhibits rapid initial descent, its aggressive adaptation combined with a lack of population diversity leads to premature convergence into local optima on tasks like MRPC.

Regarding generalization, however, \Cref{table:CR} reveals that SaES-ID underperforms BBT on test metrics. 
While searching directly in the prompt space theoretically offers a higher performance upper bound, it is more susceptible to fitting noise in the few-shot setting. 
To address this, we propose a specialized regularization term in the next section. 
By explicitly modifying the optimization objective, this approach aims to mitigate such overfitting while simultaneously reshaping the landscape to facilitate (1+1)-ES-ID in escaping local optima.
\section{Confidence-Based Regularization} \label{sec:conf_regular}
\subsection{Motivation and Formulation}
  In standard continuous prompt search, model predictions are typically determined solely by the relative magnitudes of logits within the verbalizer set $\mathcal{W}$. Consequently, an optimized prompt might achieve a low cross-entropy loss by distinguishing between candidate verbalizers, yet the resulting prediction may still exhibit low confidence when evaluated against the entire vocabulary $\mathcal{V}$. \revision{This mismatch motivates us to encourage greater probability mass on the verbalizer set, which may help reduce overfitting in few-shot prompt search.}
  
  To bridge this gap, we introduce Confidence-based Regularization (CR). The regularized loss $\ell_{\text{CR}}$ is defined as:
  \begin{equation}\label{eq:CR}
    \ell_{\text{CR}}(\bm{a}, z) = \ell_{\text{CE}}(\bm{a}, z) - \beta \cdot \log \left( \frac{\sum_{u \in \mathcal{W}} \exp(a_u)}{\sum_{v \in \mathcal{V}} \exp(a_v)} \right)
  \end{equation}
  \revision{where $\ell_{\text{CE}}$ is the cross-entropy loss defined in \cref{eq:ce}, and $\beta \in \mathbb{R}_+$ controls the regularization strength. The ratio inside the logarithm is the total probability mass assigned to $\mathcal{W}$ under the full-vocabulary softmax distribution. The regularizer penalizes low probability mass on the verbalizer set.} 

  Building upon this formulation, we incorporate $\ell_{\text{CR}}$ into our ES-ID method, yielding the variants (1+1)-ES-ID-CR and SaES-ID-CR (collectively ES-ID-CR). 
  \revision{Minimizing $\ell_{\text{CR}}$ in the full prompt space encourages greater probability mass on the verbalizer set while retaining the label-discrimination objective. Our experiments show that incorporating CR improves generalization performance on downstream tasks.}
\begin{table}[!t]
  \centering
  \caption{Performance comparison on seven GLUE tasks. For our proposed projection-free methods with the CR term, results are reported using the optimal $\beta$. ``Time/FE (s)'' represents the average execution time per function evaluation. The results marked with $^\dag$ are cited directly from \protect\cite{diao2022black}. Subscripts denote standard deviations.}
  \label{table:CR}
  \footnotesize 
  \setlength{\tabcolsep}{3pt}  
  \begin{adjustbox}{max width=\linewidth}
\begin{tabular}{ l rrrrrrrr}
    \toprule
    \multirow{2}{*}{\textbf{Algorithm}} & \textbf{MRPC} & \textbf{MNLI} & \textbf{QQP} & \textbf{SST-2} & \textbf{RTE} & \textbf{CoLA} & \textbf{QNLI} & \textbf{Time/FE} \\
     & F1 &  Acc & F1 & Acc & Acc & MCC & Acc & s \\
    \midrule    
    ManualPrompt & $77.69\phantom{_{0.00}}$ & $40.80\phantom{_{0.00}}$ & $29.37\phantom{_{0.00}}$ & $78.10\phantom{_{0.00}}$ & $50.18\phantom{_{0.00}}$ & $0.79\phantom{_{0.00}}$ & $48.70\phantom{_{0.00}}$ & - \\
    ICL          & $48.13_{12.0}$ & $47.10_{3.14}$ & $45.73_{7.01}$ & $69.36_{6.22}$ & $54.15_{2.79}$ & $-0.54_{2.67}$ & $52.06_{1.24}$ & - \\
    BDPL$^\dag$ & $78.10_{3.70}$ & $42.50_{1.80}$ & \bm{$56.40_{1.90}$} & $87.60_{2.10}$ & $53.50_{0.90}$ & $4.60_{1.20}$ & $53.10_{1.10}$ & - \\
    RLPrompt$^\dag$ & $68.90_{2.10}$ & $42.80_{3.20}$ & $53.70_{2.20}$ & $88.40_{1.90}$ & $51.80_{1.80}$ & $5.00_{1.10}$ & $52.60_{1.40}$ & - \\
    BBT & $77.33_{0.00}$ & $46.48_{1.16}$ & $51.09_{2.38}$ & $88.03_{2.02}$ & $54.30_{1.66}$ & $3.94_{1.42}$ & $54.30_{1.23}$ & 0.1442 \\
    \midrule
    ZOSGD & $77.69_{0.73}$ & $46.42_{1.28}$ & $46.16_{4.08}$ & $86.74_{0.41}$ & $53.65_{1.47}$ & $-0.81_{1.13}$ & $59.00_{1.42}$ & 0.1498 \\
    ZOSGD-CR (ours) & $79.37_{1.60}$ & $46.60_{1.38}$ & $54.04_{1.11}$ & $89.29_{1.07}$ & $54.66_{2.59}$ & $3.68_{2.00}$ & $59.48_{1.85}$ & 0.1499 \\
    \midrule
    (1+1)-ES & $77.33_{0.00}$ & $46.42_{1.28}$ & $49.52_{2.84}$ & $86.95_{0.71}$ & $52.92_{0.74}$ & $4.67_{2.13}$ & $54.42_{1.13}$ & 0.1451 \\
    (1+1)-ES-ID (ours) & $77.33_{0.00}$ & $46.42_{1.28}$ & $47.74_{5.02}$ & $86.95_{2.06}$ & $52.71_{1.44}$ & $3.71_{1.15}$ & $52.68_{1.33}$ & 0.1464 \\
    (1+1)-ES-ID-CR (ours) & \bm{$81.50_{0.10}$} & \underline{$49.04_{1.43}$} & \underline{$54.37_{0.84}$} & \underline{$89.66_{0.98}$} & \underline{$56.10_{1.16}$} & \bm{$8.32_{1.81}$} & \underline{$59.80_{2.86}$} & 0.1501 \\
    \midrule
    SaES & $77.33_{0.00}$ & $46.54_{1.05}$ & $51.62_{1.30}$ & $87.41_{1.38}$ & $52.92_{0.78}$ & \underline{$5.20_{2.60}$} & $53.84_{1.06}$ & 0.1492 \\
    SaES-ID (ours) & $77.33_{0.00}$ & $46.42_{1.28}$ & $50.93_{3.51}$ & $86.83_{1.94}$ & $53.86_{1.70}$ & $4.56_{2.88}$ & $52.88_{0.89}$ & 0.1503 \\
    SaES-ID-CR (ours) & \underline{$81.36_{0.12}$} & \bm{$52.44_{1.09}$} & $54.05_{1.07}$ & \bm{$90.37_{1.20}$} & \bm{$58.56_{2.82}$} & $4.75_{1.73}$ & \bm{$59.86_{1.37}$} & 0.1529 \\
    \bottomrule
  \end{tabular}
\end{adjustbox}
\end{table}
\subsection{Main Results}\label{sec:main-results}
\revision{\par\noindent\textit{Experimental Setup.}\enspace We follow the general experimental setup in \Cref{sec:subspace_issue}; the CR-specific hyperparameters are given below.}

\revision{\par\noindent\textit{Hyperparameters.}\enspace We fix the ID-aware adaptation dimension at $\tilde{d}=500$. For the CR variants, we search initial step sizes in $\{1,10^{-1},10^{-2},10^{-3},10^{-4}\}$ and regularization coefficients in $\{0.01,0.1,1,10,100\}$. On RoBERTa-Large, the initial step sizes are $10^{-4}$ for (1+1)-ES-ID-CR and $10^{-2}$ for SaES-ID-CR; on OPT-6.7B, both variants use an initial step size of $1$.}

\revision{We compare against ManualPrompt, in-context learning~\cite{brown2020language}, BBT~\cite{sun2022black}, RLPrompt~\cite{deng2022rlprompt}, BDPL~\cite{diao2022black}, and ZOSGD~\cite{ghadimi2013stochastic}. ZOSGD-CR applies our regularizer to ZOSGD.}
\Cref{table:CR} presents the performance and average runtime of our projection-free methods compared to these baselines.

As illustrated, our proposed ES-ID-CR variants consistently outperform the baselines across the majority of tasks. 
Compared with ManualPrompt and ICL, ES-ID-CR demonstrates remarkably stable and superior performance, whereas these baselines exhibit high variance and suffer substantial performance drops on challenging datasets like CoLA. 
Furthermore, our approach achieves higher accuracy than few-shot optimization baselines such as BBT, BDPL, and ZOSGD. 
This superiority indicates that by searching directly in the full space coupled with confidence regularization, ES-ID-CR effectively mitigates the overfitting inherent in few-shot scenarios, yielding robust generalization. 

\revision{Among our methods,} SaES-ID-CR achieves more substantial gains on MNLI and RTE compared to (1+1)-ES-ID-CR, benefiting from the robustness of population-based search.

Further analysis demonstrates that ES-ID-CR outperforms both standard ESs and ES-ID variants. 
Specifically, the ID-aware adaptation enables efficient convergence within the full prompt space, while CR serves as a critical regularizer to enhance generalization. 
To evaluate the universality of this regularization, we extend our method to ZOSGD. 
Results show that ZOSGD-CR achieves substantial improvement over standard ZOSGD, confirming CR as an effective, plug-and-play module applicable to other gradient-free optimizers. 
However, ZOSGD-CR still underperforms compared to ES-ID-CR. 
This gap arises because ZOSGD relies on pre-defined learning rate schedules rather than dimension-dependent adaptive mechanisms, fundamentally limiting its ability to exploit the intrinsic dimension for accelerated convergence. 

\revision{Finally, regarding computational efficiency, the `Time/FE' results show that our projection-free methods incur negligible overhead compared to BBT, maintaining competitive training speed despite searching in the full prompt space.}
\begin{table}[!t]
  \centering
  \caption{Sensitivity analysis of the CR coefficient $\beta$. We report the mean and standard deviation of performance over 5 runs.}
  \label{table:beta}
  \footnotesize 
  \setlength{\tabcolsep}{3pt} 
  \begin{adjustbox}{max width=\linewidth}
\begin{tabular}{ l rrrrrrr}
      \toprule
      \textbf{Algorithm} & \textbf{MRPC} & \textbf{MNLI} & \textbf{QQP} & \textbf{SST-2} & \textbf{RTE} & \textbf{CoLA} & \textbf{QNLI} \\
      \midrule
      (1+1)-ES-ID-CR$_{\beta=0.01}$ & \e{5}$77.33_{0.00}$ & \e{5}$46.42_{1.28}$ & \e{5}$38.59_{5.92}$ & \e{15}$86.88_{1.48}$ & \e{15}$54.15_{1.60}$ & \e{5}$-0.73_{1.78}$ & \e{45}$59.80_{2.86}$ \\
      (1+1)-ES-ID-CR$_{\beta=0.1}$  & \e{5}$77.33_{0.00}$ & \e{10}$46.50_{1.22}$ & \e{10}$40.84_{5.50}$ & \e{30}$88.65_{1.49}$ & \e{10}$53.72_{1.76}$ & \e{10}$0.74_{2.29}$ & \e{35}$58.08_{2.65}$ \\
      (1+1)-ES-ID-CR$_{\beta=1}$    & \e{15}$77.76_{0.64}$ & \e{45}$49.04_{1.43}$ & \e{30}$53.32_{1.39}$ & \e{45}$89.66_{0.98}$ & \e{45}$56.10_{1.16}$ & \e{45}$8.32_{1.81}$ & \e{15}$53.24_{2.46}$ \\
      (1+1)-ES-ID-CR$_{\beta=10}$   & \e{45}$81.50_{0.10}$ & \e{40}$48.72_{1.18}$ & \e{40}$54.33_{0.74}$ & \e{10}$79.50_{2.05}$ & \e{25}$54.73_{1.42}$ & \e{35}$6.16_{1.79}$ & \e{10}$52.60_{1.81}$ \\
      (1+1)-ES-ID-CR$_{\beta=100}$  & \e{35}$79.71_{1.95}$ & \e{35}$48.62_{1.12}$ & \e{45}$54.37_{0.84}$ & \e{5}$78.03_{1.14}$ & \e{15}$53.79_{0.65}$ & \e{30}$5.93_{1.48}$ & \e{5}$51.00_{1.72}$ \\
      \midrule
      SaES-ID-CR$_{\beta=0.01}$ & \s{5}$77.33_{0.00}$ & \s{5}$46.42_{1.28}$ & \s{5}$48.52_{3.96}$ & \s{25}$88.42_{2.04}$ & \s{30}$54.22_{1.92}$ & \s{10}$2.57_{1.87}$ & \s{45}$59.86_{1.37}$ \\
      SaES-ID-CR$_{\beta=0.1}$  & \s{5}$77.33_{0.00}$ & \s{15}$47.06_{0.37}$ & \s{35}$52.83_{1.17}$ & \s{35}$89.24_{1.45}$ & \s{25}$54.15_{1.43}$ & \s{10}$2.43_{2.47}$ & \s{40}$59.36_{1.32}$ \\
      SaES-ID-CR$_{\beta=1}$    & \s{20}$78.58_{1.19}$ & \s{40}$52.04_{1.14}$ & \s{45}$54.05_{1.07}$ & \s{45}$90.37_{1.20}$ & \s{45}$58.56_{2.82}$ & \s{30}$4.75_{1.73}$ & \s{15}$52.76_{1.84}$ \\
      SaES-ID-CR$_{\beta=10}$   & \s{45}$81.36_{0.12}$ & \s{30}$51.52_{1.10}$ & \s{20}$51.55_{5.51}$ & \s{10}$82.64_{5.38}$ & \s{15}$54.01_{0.49}$ & \s{25}$4.63_{3.53}$ & \s{10}$51.04_{1.32}$ \\
      SaES-ID-CR$_{\beta=100}$  & \s{40}$81.20_{0.26}$ & \s{45}$52.44_{1.09}$ & \s{15}$50.54_{7.54}$ & \s{5}$77.80_{0.80}$ & \s{15}$53.79_{0.51}$ & \s{20}$3.15_{1.63}$ & \s{5}$50.52_{1.34}$ \\
      \bottomrule
  \end{tabular}
\end{adjustbox}
\end{table}
\subsection{Sensitivity Analysis}\label{sec:sensitive}
\revision{We analyze sensitivity to the intrinsic dimension $\tilde{d}$ and regularization coefficient $\beta$.}
\par
\begin{figure}[!b] 
  \centering
  \includegraphics[width=1\columnwidth]{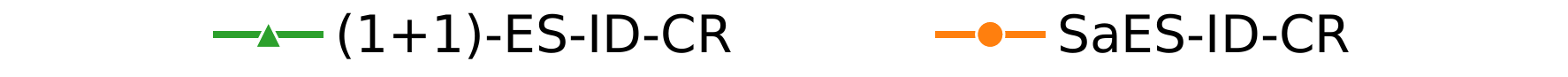} \\
  \vspace{1pt}
  \begin{subfigure}[b]{0.32\columnwidth}
    \centering
    \includegraphics[width=\linewidth]{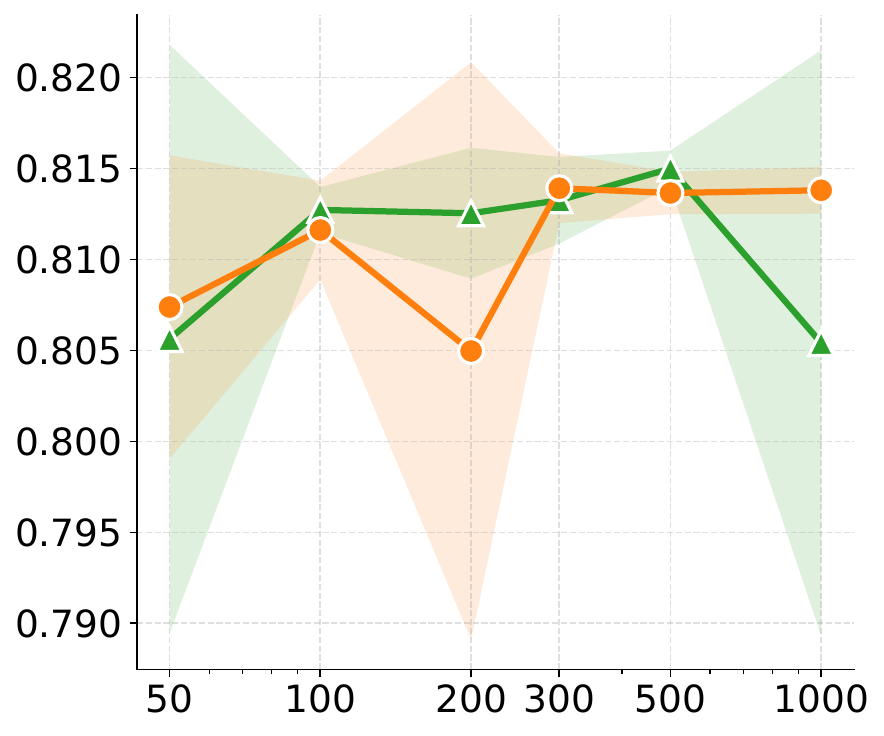}
    \caption{\scriptsize MRPC}
  \end{subfigure}
  \hfill
  \begin{subfigure}[b]{0.32\columnwidth}
    \centering
    \includegraphics[width=\linewidth]{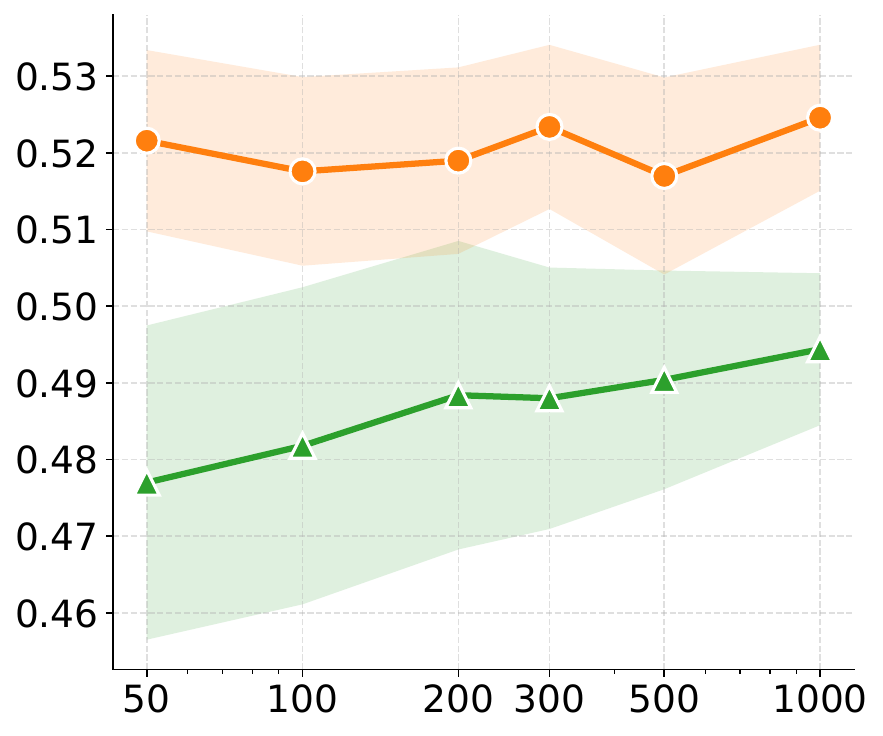}
    \caption{\scriptsize MNLI}
  \end{subfigure}
  \hfill
  \begin{subfigure}[b]{0.32\columnwidth}
    \centering
    \includegraphics[width=\linewidth]{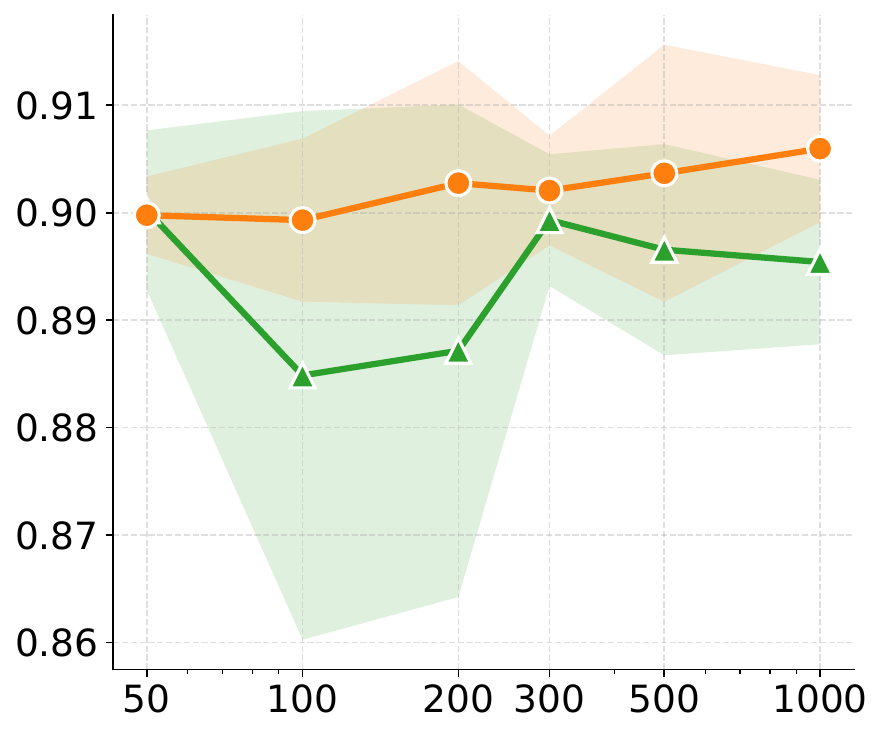}
    \caption{\scriptsize SST-2}
  \end{subfigure}
  \caption{Sensitivity analysis of intrinsic dimension $\tilde{d}$. We report the mean and standard deviation of performance over 5 runs.}
  \label{fig:differ_id}
\end{figure}

\noindent\textit{Robustness to Intrinsic Dimension $\tilde{d}$.}\enspace 
\revision{To assess the sensitivity of the configured intrinsic dimension, we evaluate ES-ID-CR with fixed $\beta$ over $\tilde{d}\in\{50,100,200,300,500,1000\}$. As illustrated in \Cref{fig:differ_id}, test performance stabilizes for the tested values of $\tilde{d}$ above the empirical upper bound ($\approx250$) identified in \Cref{sec:measureID}. Following BBT's dimension setting, we use $\tilde{d}=500$ for all tasks in our main experiments, without task-specific tuning.}
\par\noindent\textit{Impact of Regularization Coefficient $\beta$.}\enspace
With $\tilde{d}$ fixed at $500$, we investigate the regularization coefficient $\beta$ across a geometric progression $\{0.01, 0.1, 1, 10, 100\}$. 
\revision{Mathematically, $\beta$ serves as a weighting coefficient that governs the trade-off between the standard cross-entropy loss and the CR term.} 
\revision{A smaller $\beta$ reduces the weight of the probability-mass penalty relative to the label-discrimination loss, whereas a larger $\beta$ places greater emphasis on increasing the full-vocabulary probability mass assigned to the verbalizer set.}
\revision{As shown in \Cref{table:beta}, the parameter sensitivity exhibits a remarkably consistent pattern.} 
\revision{For any given task, the values yielding near-optimal performance tend to congregate within a specific magnitude regime rather than appearing as discrete, scattered peaks.} 
\revision{This smooth sensitivity landscape implies that determining $\beta$ does not necessitate computationally expensive fine-grained tuning. 
Instead, a coarse-grained search is sufficient to identify the effective regime, identifying whether a task requires dominance by cross-entropy ($\beta < 1$), a balanced objective ($\beta \approx 1$), or strong regularization ($\beta > 1$).}
\begin{table}[!t] 
  \centering
  \caption{Performance comparison on GLUE tasks using the causal OPT-6.7B model.}
  \label{tab:opt}
  \small
  \setlength{\tabcolsep}{8pt} 
  \begin{adjustbox}{max width=\linewidth}
\begin{tabular}{lccc}
    \toprule
    \textbf{Task} & \textbf{BBT} & \textbf{(1+1)-ES-ID-CR} & \textbf{SaES-ID-CR} \\
    \midrule
    MRPC  & $57.33_{8.04}$ & \underline{$74.00_{9.47}$} & \bm{$80.18_{2.26}$ }\\
    MNLI  & \underline{$32.92_{1.33}$} & $30.70_{1.16}$ & \bm{$33.54_{2.40}$} \\
    QQP   & \underline{}{$50.41_{3.28}$} & $48.02_{6.55}$ & \bm{$54.77_{3.22}$} \\
    SST-2 & \underline{$81.22_{3.15}$} &  $71.88_{3.86}$ & \bm{$86.17_{1.84}$} \\
    RTE   & \underline{$56.53_{1.75}$} & $54.95_{2.78}$ & \bm{$57.76_{1.66}$} \\
    CoLA  & \underline{$10.30_{0.94}$} & \bm{$11.42_{4.43}$} & $8.40_{0.60}$ \\
    QNLI  & \underline{$58.22_{1.58}$} & $56.30_{0.94}$ & \bm{$58.46_{0.35}$} \\
    \bottomrule
  \end{tabular}
\end{adjustbox}
\end{table}
\subsection{Generality across Model Architectures}
\label{sec:opt_experiment}
\revision{To verify generality beyond masked LMs, we evaluate our method on the causal LM OPT-6.7B}~\cite{zhang2022opt}. \revision{Despite the challenges of large-scale causal models (e.g., expanded parameter space), SaES-ID-CR outperforms BBT on 6/7 tasks (\Cref{tab:opt}), notably boosting MRPC from 57.33 to 80.18 and SST-2 from 81.22 to 86.17.} \revision{Conversely, (1+1)-ES-ID-CR is less consistent, highlighting that SaES's population diversity is essential for stable optimization in larger search spaces.} These results confirm our method's cross-architecture effectiveness.
\section{Related Work}
\subsection{Intrinsic Dimensions of PLMs}
\noindent\revision{Intrinsic dimension characterizes search-space structure and affects search efficiency.}
\revision{Recent studies measure it in neural-network parameter spaces}~\cite{li2018measuring}.
\revision{On common NLP tasks, PLM parameter spaces can have intrinsic dimensions substantially below their ambient dimensions}~\cite{aghajanyan2020intrinsic}.
\revision{Existing prompt-search methods assume low intrinsic dimension and adopt the subspace projection technique used in parameter-space studies}~\cite{sun2022black,sun2022bbtv2}.
\revision{However, this assumption remains unverified in prompt space.}
\subsection{Black-Box Prompt Search}
Prompt search is inherently black-box as PLMs expose only input-output behavior.
\revision{Evolution Strategies suit this setting, but complex variants such as CMA-ES are computationally prohibitive in high-dimensional prompt spaces.}
\revision{Recent methods therefore use subspace projection to make optimization feasible}~\cite{zhang2024subspace}.
\revision{Beyond ES, Zeroth-Order SGD~\cite{zhan2024unlocking} provides efficient gradient-free continuous prompt search.}
\revision{For discrete prompts, reinforcement-learning methods~\cite{deng2022rlprompt,diao2022black} achieve strong performance by treating prompt tokens as actions or weak classifiers.}
\section{Conclusion}
\revision{In this work, we propose a projection-free method for black-box prompt search.} 
\revision{We measure the intrinsic dimension of the prompt space, observing it exhibits an empirical upper bound independent of prompt length.} 
\revision{Guided by this, we design an ID-aware adaptation mechanism enabling efficient search in the full prompt space, bypassing the misalignment of random projections.} 
\revision{To address the generalization gap, we incorporate confidence-based regularization to encourage greater probability mass on the verbalizer set.} 
\revision{Extensive experiments demonstrate that our approach outperforms baselines across GLUE tasks and generalizes robustly to large-scale causal architectures.}
\begin{credits}
\subsubsection{Disclosure of Interests.}
\revision{The authors have no competing interests to declare that are relevant to the content of this article.}
\end{credits}
\bibliographystyle{splncs04}
\bibliography{cy}
\clearpage
\appendix
\renewcommand{\theHsection}{appendix.\arabic{section}}
\crefalias{section}{appendix}
\crefalias{subsection}{appendix}
\crefname{appendix}{Appendix}{Appendices}
\renewcommand{\thesection}{Appendix \arabic{section}}
\renewcommand{\thesubsection}{\thesection.\arabic{subsection}}
\section{Supplementary Baseline Experiments}\label{appx:1}
For ManualPrompt, we use the handcrafted templates in \Cref{table:template-roberta}.
For ICL, we prepend one randomly sampled training example per label as a demonstration.

\subsection{Additional Baseline Experimental Results}
\label{app:additional_baseline_results}

In this section, we provide supplementary experimental results for the ZOSGD-CR, BDPL, and RLPrompt baselines as presented in Table \ref{tab:additional_baselines}.

\paragraph{ZOSGD-CR Sensitivity Analysis}
In the main text, we reported the performance of ZOSGD-CR using the optimal hyperparameter configuration. Here, we investigate the impact of the regularization strength. We fixed the base learning rate at $1\text{e-}2$ and the sampling radius at $1\text{e-}4$, while presenting the complete results across varying regularization coefficients $\beta \in \{0.01, 0.1, 1, 10, 100\}$. As evidenced in Table \ref{tab:additional_baselines}, the method exhibits a consistent preference for $\beta \approx 1$ across most tasks.

\begin{table}[t]
    \caption{Additional experimental results for ZOSGD-CR, BDPL, and RLPrompt. We report the mean and standard deviation over 5 independent runs.}
    \label{tab:additional_baselines}
    \centering
    \begin{adjustbox}{max width=\linewidth}
\begin{tabular}{l rrrrrrr}
    \toprule
    \textbf{Algorithm} & \textbf{MRPC} & \textbf{MNLI} & \textbf{QQP} & \textbf{SST-2} & \textbf{RTE} & \textbf{CoLA} & \textbf{QNLI} \\
    \midrule
    ZOSGD-CR$_{\beta=0.01}$ & $77.33_{0.00}$ & $46.42_{1.28}$ & $44.74_{4.27}$ & $86.63_{0.40}$ & $53.65_{1.24}$ & $-0.25_{1.48}$ & $58.58_{1.90}$ \\
    ZOSGD-CR$_{\beta=0.1}$  & $77.58_{0.51}$ & $46.54_{1.09}$ & $48.47_{2.54}$ & $87.68_{0.52}$ & $54.66_{2.59}$ & $1.49_{2.41}$ & $59.48_{1.85}$ \\
    ZOSGD-CR$_{\beta=1}$    & $79.37_{1.60}$ & $46.42_{1.28}$ & $54.04_{1.11}$ & $89.29_{1.07}$ & $54.30_{0.54}$ & $3.68_{2.00}$ & $51.66_{1.01}$ \\
    ZOSGD-CR$_{\beta=10}$   & $77.95_{1.23}$ & $46.60_{1.38}$ & $48.63_{7.77}$ & $84.61_{2.17}$ & $53.86_{0.27}$ & $1.91_{1.64}$ & $50.46_{1.38}$ \\
    ZOSGD-CR$_{\beta=100}$  & $77.46_{0.27}$ & $46.42_{1.28}$ & $48.23_{8.85}$ & $82.39_{4.15}$ & $53.57_{0.43}$ & $3.45_{3.85}$ & $51.18_{1.76}$ \\
    \midrule
    BDPL (Ours)             & $68.63_{4.95}$ & $25.76_{1.97}$ & $52.40_{1.21}$ & $83.85_{1.57}$ & $52.78_{2.38}$ & $1.95_{4.20}$ & $54.35_{1.40}$ \\
    RLPrompt (Ours)         & $47.53_{1.18}$ & $48.73_{2.23}$ & $49.11_{2.00}$ & $84.58_{3.12}$ & $56.61_{2.27}$ & $-1.66_{2.63}$ & $51.19_{1.20}$ \\
    \bottomrule
    \end{tabular}
\end{adjustbox}
\end{table}
\paragraph{Baseline Reproduction and Analysis} 
To ensure a rigorous comparative study, we reproduced the discrete optimization baselines, BDPL \cite{diao2022black} and RLPrompt \cite{deng2022rlprompt}, strictly adhering to the hyperparameter configurations and computational budgets recommended in their original literature. 

We observe that the reproduction results under our unified 16-shot protocol are generally lower than the best results reported in the original papers. We attribute this to two technical factors inherent to the unified evaluation setting:
\begin{itemize}
    \item \textbf{Metric-Reward Misalignment in RLPrompt:} The default reward function in the official RLPrompt implementation is designed to optimize prediction accuracy. This creates an objective mismatch for tasks evaluated on disjoint metrics such as F1 score (MRPC, QQP) or MCC (CoLA). Without task-specific reward engineering (which introduces additional manual intervention), the policy learning remains suboptimal for these specific metrics.
    \item \textbf{Sensitivity in Discrete Search:} Methods like BDPL are known to be sensitive to seed variations in few-shot scenarios. Our reproduction reflects the average performance under a standard, unrefined execution without extensive random seed searching.
\end{itemize}

\textit{Note on Fairness:} We observed that the BDPL literature \cite{diao2022black} reported significantly higher performance for RLPrompt compared to our local reproduction, although the specific implementation details yielding these results were not fully disclosed. Given the high overlap between our evaluation tasks and those in \cite{diao2022black}, we adopt the most rigorous comparison strategy: in the main text (Table 2), we cite the reported results from \cite{diao2022black} for both their proposed method (BDPL) and the RLPrompt baseline. The local reproduction results in Table \ref{tab:additional_baselines} are provided solely to illustrate the methods' behavior under a constrained, out-of-the-box experimental setting.

\section{Backbone Models and Templates}
\label{appx:2}

We conduct experiments on two distinct pre-trained language models to verify the generality of our framework: RoBERTa-Large (355M parameters), representative of masked language models, and OPT-6.7B (6.7B parameters), representative of large-scale causal language models.

For each task, we assign specific manual templates and verbalizers. We primarily adopt the templates from BDPL~\cite{diao2022black} with minor modifications to fit the respective model architectures. \Cref{table:template-roberta} and \Cref{table:template-opt} provide the detailed configurations for RoBERTa-Large and OPT-6.7B, respectively.

\begin{table}[!ht]
    \caption{Templates and label descriptions for the mask-filling RoBERTa-Large model.}
    \label{table:template-roberta}
    \centering
    \small
    \begin{adjustbox}{max width=\linewidth}
\begin{tabular}{l l l}
    \toprule
    \textbf{Dataset} & \textbf{Template} & \textbf{Verbalizers} \\
    \midrule
    MRPC    & $s_1$? [MASK], $s_2$.& yes / no         \\
    MNLI    & $s_1$? [MASK],$s_2$.      & yes / maybe / no \\
    QQP     & $s_1$ [MASK],$s_2$.       & yes / no         \\
    SST-2    & $s_1$. It was [MASK].             & great / terrible \\
    RTE     & $s_1$? [MASK],$s_2$.      & yes / no         \\
    CoLA    & $s_1$ Correct? [MASK].            & yes / no         \\
    QNLI    & $s_1$? [MASK],$s_2$.      & yes / no         \\
    \bottomrule
    \end{tabular}
\end{adjustbox}
\end{table}
\begin{table}[!ht]
  \centering
  \caption{Templates and label descriptions for the causal OPT-6.7B model.}
  \label{table:template-opt}
  \footnotesize
  \begin{tabularx}{\columnwidth}{@{} l X l @{}}
    \toprule
    \textbf{Dataset} & \textbf{Template} & \textbf{Verbalizers} \\
    \midrule
    MRPC & input: sentence one: $s_1$ sentence two: $s_2$ equivalent? \textbackslash n output: & yes / no \\ \addlinespace[2pt]
    
    MNLI & input: sentence one: $s_1$ sentence two: $s_2$ entailment? \textbackslash n output: & yes / maybe / no \\ \addlinespace[2pt]
    
    QQP  & input: sentence one: $s_1$ sentence two: $s_2$ equivalent? \textbackslash n output: & yes / no \\ \addlinespace[2pt]
    
    SST-2 & input: $s_1$ It was \textbackslash n output: & great / terrible \\ \addlinespace[2pt]
    
    RTE  & input: sentence one: $s_1$ sentence two: $s_2$ entailment? \textbackslash n output: & yes / no \\ \addlinespace[2pt]
    
    CoLA & input: $s_1$ correct? \textbackslash n output: & yes / no \\ \addlinespace[2pt]
    
    QNLI & input: sentence one: $s_1$ sentence two: $s_2$ entailment? \textbackslash n output: & yes / no \\
    \bottomrule
  \end{tabularx}
\end{table}
\FloatBarrier
\section{Additional Experimental Details}\label{appx:3}

\subsection{Design of the Random Projection Comparison}
\label{app:subspace_analysis_settings}
The setup is described in \Cref{sec:subspace_issue}. At the default prompt length, full-space CMA-ES would store roughly 2.6 billion covariance entries. Shortening the prompt increases the ratio of subspace to ambient dimensions from less than 1\% to about 10\%, giving the projected search a larger fraction of the available dimensions.

\subsection{Step-Size Alignment for Fair Comparison}\label{appx:3.3}

For the comparison without CR in \Cref{sec:id}, we align the expected squared norms of the initial updates in the full prompt space.

Let $\Delta_{\text{ES}}$ and $\Delta_{\text{BBT}}$ denote the update vectors in the full space for ES and BBT, respectively. For ES, the update is given by $\Delta_{\text{ES}} = \sigma_{\text{ES}} \cdot \mathbf{u}$, where $\mathbf{u} \sim \mathcal{N}(\mathbf{0}, \mathbf{I}_d)$. The expected squared Euclidean norm is:
\begin{equation}\label{eq:appendix-es-energy}
    \mathbb{E}[\|\Delta_{\text{ES}}\|^2] = \sigma_{\text{ES}}^2 \cdot \mathbb{E}[\|\mathbf{u}\|^2] = \sigma_{\text{ES}}^2 \cdot d.
\end{equation}
For BBT, the update is $\Delta_{\text{BBT}} = \mathbf{A}(\sigma_{\text{BBT}} \cdot \mathbf{z})$, where $\mathbf{z} \sim \mathcal{N}(\mathbf{0}, \mathbf{I}_{\tilde{d}})$ and entries of $\mathbf{A}$ are sampled from $U[-\frac{1}{\sqrt{\tilde{d}}}, \frac{1}{\sqrt{\tilde{d}}}]$. The expected squared norm is calculated as:
\begin{align}
    \mathbb{E}[\|\Delta_{\text{BBT}}\|^2] &= \sigma_{\text{BBT}}^2 \mathbb{E}_{\mathbf{A}, \mathbf{z}}[\text{Tr}(\mathbf{z}^\top \mathbf{A}^\top \mathbf{A} \mathbf{z})] \nonumber \\
    &= \sigma_{\text{BBT}}^2 \mathbb{E}_{\mathbf{A}}[\text{Tr}(\mathbf{A}\mathbf{A}^\top)] \nonumber \\
    &= \sigma_{\text{BBT}}^2 \sum_{i=1}^d \sum_{j=1}^{\tilde{d}} \mathbb{E}[A_{i,j}^2].
\end{align}
Given that $\mathbb{E}[A_{i,j}^2] = \frac{1}{3\tilde{d}}$ for the uniform distribution, we obtain:
\begin{equation}\label{eq:appendix-bbt-energy}
    \mathbb{E}[\|\Delta_{\text{BBT}}\|^2] = \sigma_{\text{BBT}}^2 \cdot (d \cdot \tilde{d}) \cdot \frac{1}{3\tilde{d}} = \sigma_{\text{BBT}}^2 \cdot \frac{d}{3}.
\end{equation}
By equating \cref{eq:appendix-es-energy,eq:appendix-bbt-energy} to match the update energy ($\mathbb{E}[\|\Delta_{\text{ES}}\|^2] =$ $\mathbb{E}[\|\Delta_{\text{BBT}}\|^2]$), we find that $\sigma_{\text{ES}}^2 \cdot d = \sigma_{\text{BBT}}^2 \cdot \frac{d}{3}$, which simplifies to:
\begin{equation}
    \sigma_{\text{ES}} = \frac{\sigma_{\text{BBT}}}{\sqrt{3}}.
\end{equation}
Consequently, with the BBT default $\sigma_{\text{BBT}}=1$~\cite{sun2022black}, we configure $\sigma^{[0]} = 1/\sqrt{3}$ for the ES variants (standard ES and ES-ID) in our comparative experiments.

This step-size alignment applies to the comparison without CR. Adding CR changes the optimization objective and thus the search trajectory, so for the CR experiments we use the initial step sizes specified in \Cref{sec:main-results}.

\section{Mechanism of Confidence Regularization}
\label{app:cr-mechanism}

To understand the generalization gap in projection-free prompt search, we analyze the model behavior using two diagnostic metrics evaluated on the test set: \textbf{Prediction Probability} ($\mathcal{P}$) and \textbf{Global Rank} ($\mathcal{R}$).

Standard prediction strategies normalize probabilities only over the restricted verbalizer set $\mathcal{W}$, masking the model's true uncertainty. To reveal this, we define Prediction Probability ($\mathcal{P}$) as the average probability assigned to the \textit{predicted} verbalizer $\hat{z}$ when normalized over the entire vocabulary $\mathcal{V}$. Formally:
\begin{equation}
    \mathcal{P} = \frac{1}{|\mathcal{D}_{\text{test}}|} \sum_{(x) \in \mathcal{D}_{\text{test}}} \frac{\exp(\mathbf{h}_{\hat{z}})}{\sum_{v \in \mathcal{V}} \exp(\mathbf{h}_v)},
\end{equation}
where $\hat{z}$ is the verbalizer with the highest logit within $\mathcal{W}$. A low $\mathcal{P}$ implies that while the model selects $\hat{z}$ from the candidates, it considers other irrelevant tokens in $\mathcal{V}$ to be more probable. Complementarily, we employ Global Rank ($\mathcal{R}$), which computes the average rank of the \textit{predicted} verbalizer's logit within the full vocabulary $\mathcal{V}$. A rank closer to 1 indicates that the predicted label is naturally prioritized by the PLM without artificial constraints.

\subsection{Analysis}

We conducted a comparative analysis between SaES-ID (without regularization) and SaES-ID-CR (with regularization), with results summarized in Table \ref{tab:cr_mechanism}.

\begin{table}[!htbp]
    \caption{Impact of Confidence Regularization on task performance and diagnostic metrics.}
    \label{tab:cr_mechanism}
    \centering
    \begin{adjustbox}{max width=\linewidth}
    \setlength{\tabcolsep}{8pt}
    \begin{tabular}{llccc}  
    \toprule
    \textbf{Task} & \textbf{Algorithm} & \textbf{Metric (\%)} & $\mathcal{P}$ & $\mathcal{R}$ \\ \midrule
    \multirow{2}{*}{MRPC} & SaES-ID    & $76.68_{0.00}$ & 0.0017 & 302.23 \\
                          & SaES-ID-CR & $81.26_{0.11}$ & 0.4710 & 16.80  \\ \midrule
    \multirow{2}{*}{MNLI} & SaES-ID    & $46.40_{1.64}$ & 0.0099 & 221.48 \\
                          & SaES-ID-CR & $51.84_{1.37}$ & 0.0496 & 19.50  \\ \midrule
    \multirow{2}{*}{QQP}  & SaES-ID    & $49.51_{3.21}$ & 0.0017 & 203.45 \\
                          & SaES-ID-CR & $53.28_{2.93}$ & 0.3021 & 3.25   \\ \midrule
    \multirow{2}{*}{SST-2} & SaES-ID    & $87.32_{1.72}$ & 0.0403 & 8.76   \\
                          & SaES-ID-CR & $90.25_{0.47}$ & 0.1655 & 1.58   \\ \midrule
    \multirow{2}{*}{RTE}  & SaES-ID    & $53.65_{1.74}$ & 0.0036 & 159.51 \\
                          & SaES-ID-CR & $57.47_{1.83}$ & 0.3188 & 5.32   \\ \midrule
    \multirow{2}{*}{CoLA} & SaES-ID    & $3.86_{2.18}$  & 0.0007 & 342.53 \\
                          & SaES-ID-CR & $5.84_{3.09}$  & 0.3271 & 1.33   \\ \midrule
    \multirow{2}{*}{QNLI} & SaES-ID    & $54.28_{1.49}$ & 0.0034 & 184.53 \\
                          & SaES-ID-CR & $60.10_{2.20}$ & 0.0036 & 128.52 \\ \bottomrule
    \end{tabular}
    \end{adjustbox}
\end{table}

A critical observation is that optimization without CR tends to exploit the restricted decision space. As evidenced by the extremely poor Global Rank (often exceeding hundreds) despite reasonable task metrics, standard SaES-ID finds prompts that marginally push the correct verbalizer above incorrect candidates within $\mathcal{W}$, while leaving them buried beneath unrelated functional tokens in the global distribution. These prompts effectively "game" the few-shot objective, achieving high training accuracy without establishing a robust semantic connection to the task. Such prompts are "correct for the wrong reasons" and are highly susceptible to overfitting the few-shot training examples.

Compared with SaES-ID, SaES-ID-CR assigns a higher average full-vocabulary probability to the predicted verbalizer on six of the seven tasks and improves its average global rank on all seven tasks, as shown in \Cref{tab:cr_mechanism}.

\end{document}